\documentclass[letterpaper, 10 pt, conference, twoside]{ieeeconf}
\usepackage{times}
\usepackage[pdftex]{graphicx}
\usepackage{subfig}
\usepackage{amsmath,amssymb,amsopn,amstext,amsfonts}
\usepackage{cancel}
\usepackage[space]{cite}
\usepackage{pdfsync}
\usepackage{balance}
\usepackage{xcolor}
\usepackage{color}
\usepackage{mathtools}
\usepackage{algorithm}
\usepackage{algorithmic}
\usepackage{bm}

\usepackage{diagbox}
\usepackage{float}
\usepackage{epstopdf}
\usepackage{url}
\usepackage{booktabs}
\usepackage{framed}
\usepackage{lineno}
\usepackage[linkcolor=black,citecolor=black,urlcolor=black,colorlinks=true]{hyperref}
\usepackage{verbatim}
\usepackage{subfig}
\usepackage{bm}
\usepackage{afterpage} 
\usepackage[font=small]{caption}
\makeatletter
\def\tagform@#1{\maketag@@@{\normalsize(#1)\@@italiccorr}}
\makeatother

\graphicspath{{./Figures/}}
\DeclareGraphicsExtensions{.pdf,.png,.jpg,.eps,.svg}
\IEEEoverridecommandlockouts
\vspace{1.4cm}
\title{\LARGE \bf Aerial Grasping via Maximizing Delta-Arm Workspace Utilization}
\author{Haoran Chen$^{1*}$, Weiliang Deng$^{1*}$, Biyu Ye$^{1}$, Yifan Xiong$^{1}$, Zongliang Pan$^{2}$, and Ximin Lyu$^{1\dagger}$
    \thanks{${}^{*}$ \textbf{Equal contribution}.
    ${}^{\mathbf{\dagger}}$ \textbf{Corresponding Author}.}
    \thanks{This work is supported by the National Key Research and Development Program of China (Grant No. 2023YFB4706600) and the National Natural Science Foundation of China (Grant No. 62303495).}
    \thanks{$^1$School of Intelligent Systems Engineering, Sun Yat-sen University, Guangzhou, China. $^2$Shenzhen ePropulsion Technology Limited, Shenzhen, China. (Email: lvxm6@mail.sysu.edu.cn)}
}
\begin{document}
\maketitle

\begin{abstract}
Workspace limitations restrict the operational capabilities and range of motion for systems with robotic arms.
Maximizing workspace utilization has the potential to provide better solutions for aerial manipulation tasks, increasing the system's flexibility and operational efficiency. 
In this paper, we introduce a novel planning framework for aerial grasping that maximizes workspace utilization. 
We formulate an optimization problem to optimize the aerial manipulator's trajectory, incorporating task constraints to achieve efficient manipulation. 
To address the challenge of incorporating the delta arm's non-convex workspace into optimization constraints, we leverage a Multilayer Perceptron
(MLP) to map the point positions to feasibility probabilities.
Furthermore, we employ Reversible Residual Networks (RevNet) to approximate the complex forward kinematics of the delta arm, utilizing its efficient model gradients to further eliminate workspace constraints. 
We validate our methods in simulations and real-world experiments to demonstrate their effectiveness.
\end{abstract}

\section{Introduction}
Recently, the research community has shown considerable interest in aerial manipulators that integrate robotic arms with Unmanned Aerial Vehicles (UAVs), owing to their promising performance in industrial and service applications. 
The ability to swiftly access and operate in challenging environments distinguishes these aerial systems from their fixed-base or ground-based counterparts, enabling them to handle a wider spectrum of manipulation tasks, like manufacturing, perching, and especially grasping~\cite{bauer2023autonomous,zhang2022aerial,huang2019perching}.

Previous research in aerial manipulators has primarily focused on developing precise control~\cite{kumar2024thrust},~\cite{wang2024impact} and advancing hardware design~\cite{guo2024powerful},~\cite{firouzeh2024perching} to expand application scopes and capability boundaries. 
However, the planning component in manipulation systems is of equal importance, yet has received insufficient research attention.
In this paper, we focus on grasp planning for aerial manipulators with delta arms, ensuring compliance with dynamic, kinematic, and task constraints (cf. Fig.~\ref{fig:cover}).


Two main challenges need to be addressed in this domain:  \textbf{(i) High-Dimensional Search Space}: Aerial manipulators integrate the degrees of freedom (DoFs) of UAVs and robotic arms, resulting in a significantly larger search space during motion planning \cite{deng2025wholebody}. 
\textbf{(ii) Non-Convexity of Delta Arm Workspace}: The workspace of delta arms exhibits pronounced non-convex characteristics, making it challenging to represent and formulate constraints within the planning process. 
While existing methods incorporate delta arm workspace constraints, they often approximate the entire workspace using a single convex polyhedron~\cite{cao2025motion}. 
This simplification results in the vast majority of the available workspace remaining unutilized, significantly limiting the delta arm’s end-effector range of motion.

\begin{figure*}[t]
    \centering
    \includegraphics[width=0.99\textwidth]{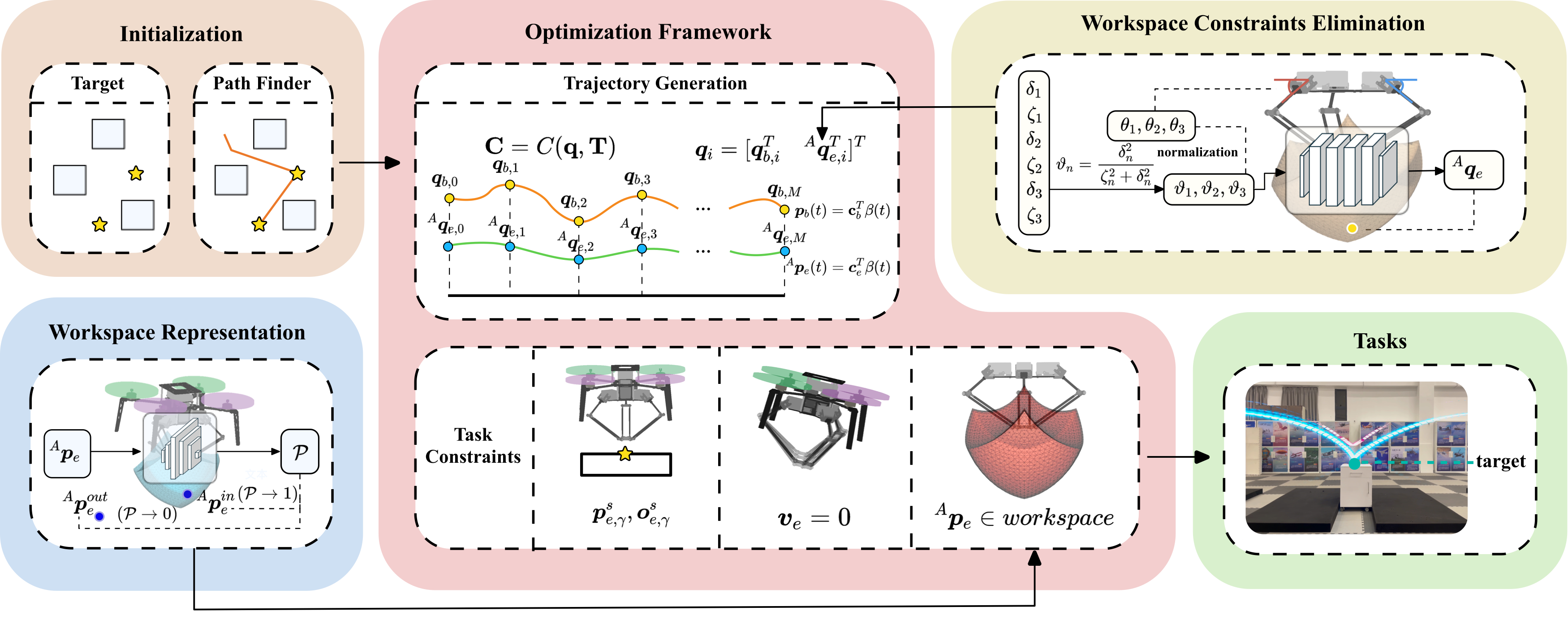}
    \caption{The overview of our proposed manipulation planning framework for aerial manipulators with delta arms, which mainly consists of four modules. 
    \textbf{(a)} The initialization utilizes a pathfinder to generate a reference path traversing all target points. 
    \textbf{(b)} The workspace representation module leverages a learning-based method to map coordinate points to feasibility possibilities. 
    \textbf{(c)} The whole-body optimization problem is formulated by continuous polymathic trajectory generation and task constraints for manipulation tasks. 
    \textbf{(d)} The workspace constraint elimination module approximates the forward kinematics of the delta arm with a learning model to calculate mapping and gradient. 
    Finally, the aerial manipulator performs successful aerial grasping in real-world experiments.}
    \label{fig:overview}
\end{figure*}

The overview of our framework is illustrated in Fig.~\ref{fig:overview}.
Our main contributions are:

\begin{figure}[t]
    \centering
    \includegraphics[width=\columnwidth]{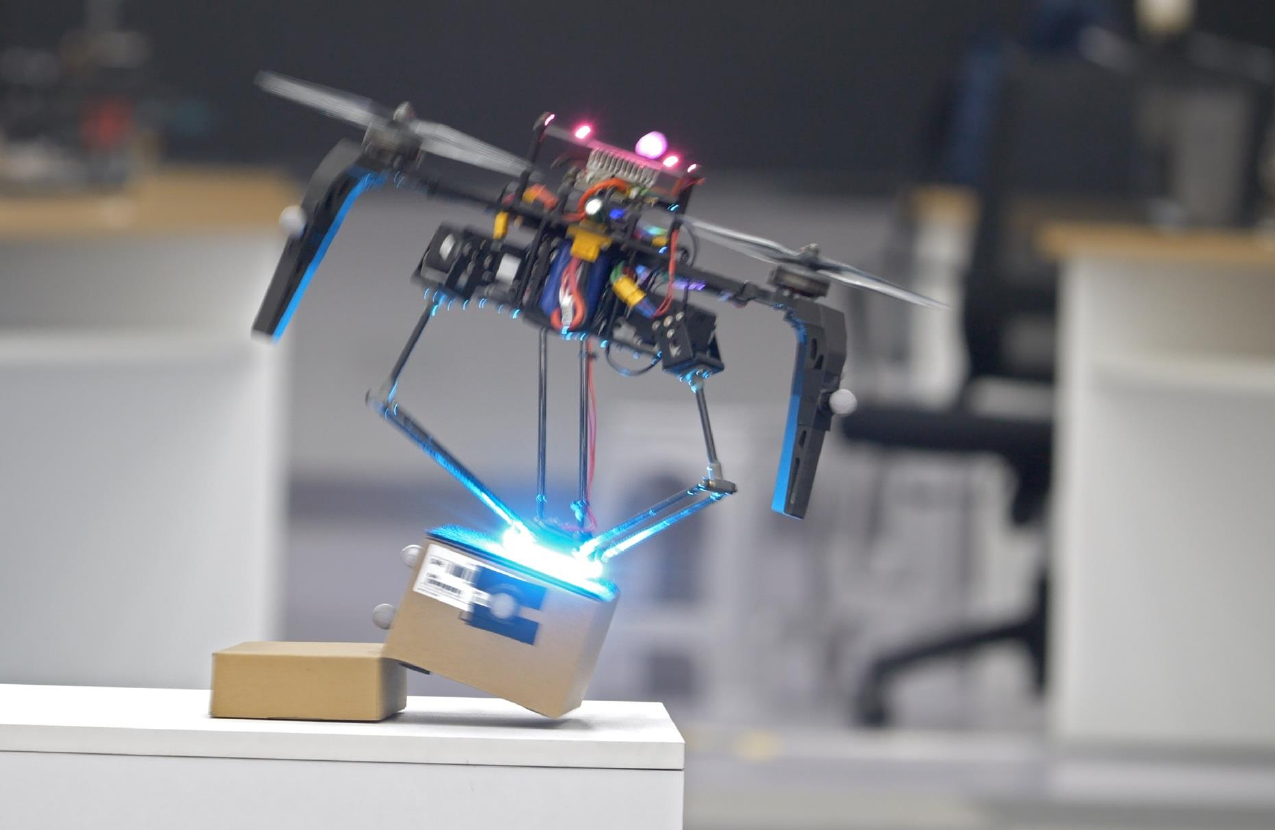}
    \caption{The aerial manipulator successfully navigates to the target point and executes high-speed grasping of the inclined object.}
    \label{fig:cover}
    \vspace{-1.0cm}
\end{figure}

\begin{itemize}
    \item A novel whole-body planning framework for aerial manipulators with delta arms that satisfies task-specific end-effector constraints and complex workspace constraints. 
    \item An MLP architecture to transform Cartesian coordinates into feasibility probabilities, providing an intuitive representation of the complex, non-convex delta workspace. 
    \item A RevNet model for accurate delta arm forward kinematics and a method that eliminates the need for explicit workspace constraints of optimization variables.
    \item Validation of our approach through extensive simulations and real-world experiments, demonstrating successful aerial grasping across multiple scenarios. 
\end{itemize}

\section{Related Works}
\subsection{Planning for Aerial Manipulators}
The planning methods for aerial manipulators can be divided into two main categories: decoupled planning and coupled planning \cite{deng2025wholebody}. Decoupling the trajectory planning of UAVs and robotic arms is a common practice in manipulation planning, as this separation significantly simplifies the planning process and allows for the straightforward application of established methods from both domains. 
For example, Dimmig \textit{et al.} \cite{dimmig2023small} and Garimella \textit{et al.}~\cite{garimella2021improving} implement a strategy where the UAV flies to a manually specified point near the target before executing predetermined grasping motions. 
Similarly, Chen \textit{et al.} ~\cite{chen2019aerial} and Ramon \textit{et al.} ~\cite{ramon2019grasp} leverage sensors, such as cameras, to detect when the UAV is sufficiently close to the target, at which point it hovers before planning the robotic arm's operational trajectory. 
Despite their simplicity, these decoupled methods, which position the UAV at rule-based predetermined locations, cannot guarantee global optimality of the complete motion sequence and also result in suboptimal execution speed due to their two-phase planning approach.

Coupled planning methods optimize UAV and robotic arm trajectories simultaneously, addressing these limitations but resulting in higher-dimensional state spaces with increased computational demands. Su \textit{et al.} \cite{su2023sequential} model the system as a virtual kinematic chain, which simplifies optimization but applies only to specific UAV configurations. Lee \textit{et al.} \cite{lee2018planning} and Tognon \textit{et al.}\cite{tognon2018controlaware} use RRT* for path searching, but the low search dimensionality
and trajectory sampling constrain UAV maneuverability~\cite{ren2022bubble}.  Cuniato \textit{et al.}~\cite{cuniato2023learning} applies reinforcement learning to develop flight strategies, achieving good results in complex tasks but performing poorly in novel, untrained scenarios. Our proposed method utilizes a coupled trajectory optimization architecture that addresses these limitations.

\subsection{Workspace Constraint Modeling}
The robotic arm's workspace plays a crucial role in trajectory optimization by preventing the inclusion of unreachable points~\cite{zacharias2007capturing}. 
Conventional feasible space constraint formulation methods are typically limited to convex polyhedral regions in three-dimensional position space, which simplifies implementation but restricts flexibility. This limitation becomes particularly problematic for delta arms, which possess inherently non-convex workspaces (cf. Fig. \ref{fig:represent} (a)), making direct integration of delta workspace constraints impractical using traditional approaches.
To address this challenge, various approximation methods have been proposed. Deng \textit{et al.} \cite{deng2025wholebody} approximate the workspace using a slender rectangular prism extending along the z-axis in the arm's coordinate system.
Although this approach preserves convexity and fully utilizes the manipulator's motion capabilities along the $z$-axis, it fails to exploit the full range of motion along the $x$ and $y$ axes, limiting overall system performance. 
Similarly, Cao \textit{et al.} \cite{cao2025motion} employ the maximum inscribed rectangular prism within the workspace, which provides a simple representation but still achieves relatively low workspace utilization.
To overcome these limitations, our method utilizes a neural network to map 3D coordinates to feasibility probabilities, enabling direct optimization within the entire non-convex workspace and enhancing the system's motion flexibility.

\section{Preliminaries}

\subsection{Coordinates Definition and Dynamic Model}

The aerial manipulator discussed in this paper consists of a quadrotor and a delta arm. We model the system using three coordinate frames: the World Frame ($\mathcal{F}_W$), the Body Frame ($\mathcal{F}_B$), and the Delta Arm Frame ($\mathcal{F}_A$). 
The positive z-axis of $\mathcal{F}_W$ points opposite to the gravity direction. 
The origin of ($\mathcal{F}_B$) is located at the system's Center of Mass (CoM). 
The x-axis of $\mathcal{F}_B$ points toward the front of the quadrotor, while the z-axis points in the direction of the total thrust generated by the four rotors. 
$\mathcal{F}_A$ is positioned on the base of the delta arm. As the base is rigidly attached to the quadrotor, the transformation between $\mathcal{F}_A$ and $\mathcal{F}_B$ can be represented by a translational transformation. 
In this work, the left superscripts of vectors indicate the frame in which the vector is expressed. For simplicity, the left superscripts of the vectors expressed in $\mathcal{F}_W$ will be omitted. 

The aerial manipulator presented in this paper has a total of 9 DoFs, with 6 DoFs from the quadrotor and 3 from the delta arm. 
Since the end-effector of the delta arm remains parallel to the base throughout the operation, it provides only 3 translational DoFs. 
In this paper, we treat the UAV and the robotic arm as two separate systems. The configuration of the UAV is defined by its position $\boldsymbol{p}_b$ and the rotation matrix $\mathbf{R}_b$, while the robotic arm's is defined by ${}^{A}\boldsymbol{p}_e$. The simplified dynamics of the aerial manipulator proposed by \cite{chen2025ndob} is written as:
\begin{equation}
\label{equ:dynamics}
    \begin{cases}
        \boldsymbol{f}=\ddot{\boldsymbol{p}}_b+g \boldsymbol{z}_{\mathcal{W}} \\ 
        \boldsymbol{\tau}=\mathbf{I} \cdot{}^{B} \dot{\boldsymbol{\omega}}_b+{}^{B} \boldsymbol{\omega}_b \times\left(\mathbf{I} \cdot{}^{B} \boldsymbol{\omega}_b\right) \\ 
        \dot{\mathbf{R}}_b = \mathbf{R}_b{}^{B}\hat{\boldsymbol{\omega}}_b \\
        {}^{A}\boldsymbol{p}_{e} = h_e(\boldsymbol{\theta})
    \end{cases}
\end{equation}
where $\boldsymbol{f}, \boldsymbol{\tau} \in \mathbb{R}^{3}$ represent the total normalized thrust and total torque of the quadrotor, respectively. $g$ is the gravitational acceleration, and $\boldsymbol{z}_{\mathcal{W}}$ is the z-axis of $\mathcal{F}_W$. $\mathbf{I}$ represents the moment of the inertia matrix, and ${ }^{B} \boldsymbol{\omega}_b$ is the body angular velocity in $\mathcal{F}_B$. ${ }^{B}\hat{\boldsymbol{\omega}}_b$ is the skew-symmetric matrix form of ${ }^{B}\boldsymbol{\omega}_b$. $\boldsymbol{\theta}=[\theta_1 \quad \theta_2 \quad \theta_3]^T$ is the three joint angles of the delta arm ($\theta_i\in[0,90], i=1,2,3$), and $h_e(\cdot):\mathbb{R}^3\to\mathbb{R}^3$ denotes the forward kinematics.

\subsection{Trajectory Representation}

In this paper, we adopt the MINCO representation~\cite{wang2022geometrically}, using a piece-wise polynomial trajectory to conduct spatial-temporal deformation. The entire trajectory for aerial manipulators with delta arms can be represented as:
\begin{equation}
\label{equ:minco_trajectory}
\begin{aligned}
    \xi_{MINCO}= & \left\{\boldsymbol{p}(t):\left[0, T_{\sigma}\right] \mapsto \mathbb{R}^6 \mid \mathbf{C}=C(\mathbf{q}, \mathbf{T}),\right. \\
    & \left.\mathbf{q} \in \mathbb{R}^{(M-1)\times 6}, \mathbf{T} \in \mathbb{R}_{>0}^M\right\}.
\end{aligned}
\end{equation}
In this representation, $\boldsymbol{p}(t)$ is a six-dimensional vector where the first three dimensions correspond to the quadrotor position and the last three dimensions describe the delta arm configuration.
$\mathbf{C}=[\mathbf{c}_1^T \cdots  \mathbf{c}_M^T]^T$ is the polynomial coefficient matrix, $\mathbf{q}=[\boldsymbol{q}_1 \cdots \boldsymbol{q}_{M-1}]^T$ the intermediate points, $\mathbf{T}=[T_1 \cdots \ T_M]^T$ the time vector, $T_{\sigma}=\sum_{i=1}^M T_i$ the total time, and $C(\mathbf{q}, \mathbf{T})$ the parameter mapping constructed from Theorem 2 in~\cite{wang2022geometrically}.

The $i^{th}$ piece trajectory $\boldsymbol{p}_i(t)$ is represented as:
\begin{equation}
\label{equ:trajectory_piece}
    \boldsymbol{p}_i(t)=\mathbf{c}_i^T \mathbf{\beta}(t), \quad \forall t \in\left[0, T_i\right]
\end{equation}
where $\mathbf{c}_i \in \mathbb{R}^{2s\times6}$ is the coefficient matrix of the $i^{th}$ piece. 
For computational convenience, we split $\mathbf{c}_i$ into $\mathbf{c}_{b,i} \in \mathbb{R}^{2s\times3}$ for the aerial manipulator and $\mathbf{c}_{e,i} \in \mathbb{R}^{2s\times3}$ for the end-effector. 
$s$ represents the order of the integrator chain. $\mathbf{\beta}(t)=[1 \quad t \quad t^2 \quad \cdots \quad t^{2s-1}]^T$ is the natural basis vector and $T_i = t_i - t_{i-1}$ is the duration of the $i^{th}$ piece.

\section{Methodologies}

\subsection{Workspace Representation}
\label{sec:workspace_represent}
In the optimization of the delta arm trajectory ${}^{A}\boldsymbol{p}_e(t)$, we need to properly design a penalty function that constrains the trajectory within the workspace while also obtaining a gradient that can effectively guide the point ${}^{A}\boldsymbol{p}_e$ on the trajectory back into the workspace when necessary.
For grasp planning, handling end-effector position and velocity in Cartesian space is more intuitive. 
However, the challenge lies in the notably non-convex nature of the delta arm's workspace, which makes it difficult to mathematically represent the workspace in a way that generates ideal optimization gradients.

\begin{figure}[t]
    \centering
    \includegraphics[width=\columnwidth]{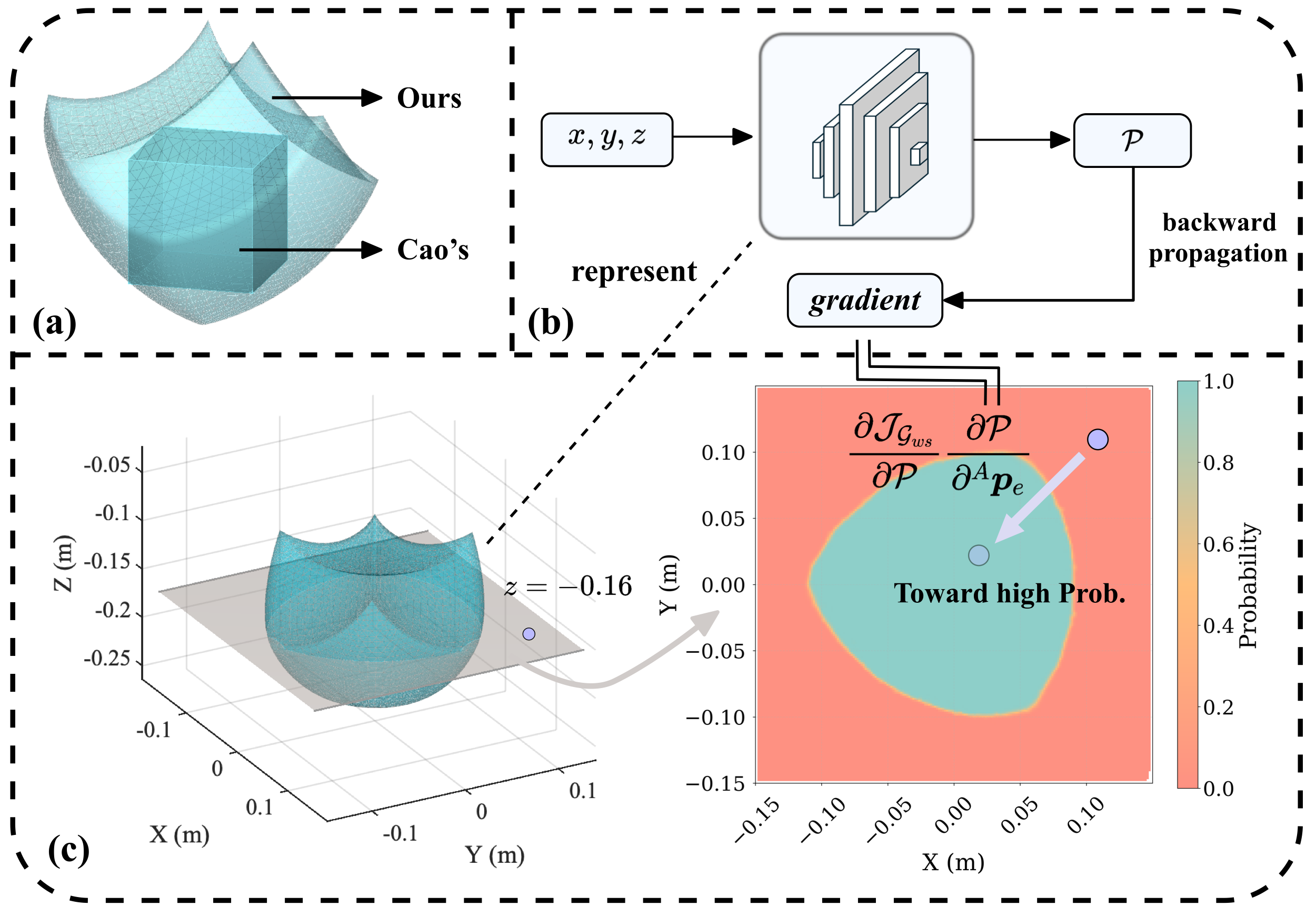}
    \caption{Workspace representation method. \textbf{(a)} Comparison of baseline representation method~\cite{cao2025motion} and ours. \textbf{(b)} The workflow of our representation model. \textbf{(c)} The process of how an optimization point is navigated to the workspace.}
    \label{fig:represent}
    \vspace{-0cm}
\end{figure}

Previous approaches employ an inscribed cube to approximate the workspace \cite{cao2025motion}, providing an efficient representation for penalty and gradient calculation.
However, this approximation significantly reduces the end-effector's feasible space during planning, limiting the aerial manipulator's manipulation flexibility.
Furthermore, the exclusion of the non-convex portion (mainly the upper portion) of the workspace forces the robotic arm to maintain a highly extended state during flight, which compromises control accuracy and introduces potential risks.
To overcome these limitations, we propose a learning-based method to represent the delta workspace, which enables maximum utilization of the entire workspace in the planning process. 
The comparison of the two methods is shown in Fig.~\ref{fig:represent} (a).

The optimization process requires determining the relative position between ${}^{A}\boldsymbol{p}_e$ and the workspace, which can be approximated as a binary classification problem using feasibility probability to distinguish points inside and outside the workspace. The problem can be modeled as:

\begin{equation}
\label{workspace_representation}
    \mathcal{P} = F_{w}({}^{A}\boldsymbol{p}_e)
\end{equation}
where $\mathcal{P}\in[0,1]$ represents the probability of a point being in the workspace, and $F_{w}(\cdot)$ is a mapping function from three-dimensional point coordinates to feasibility probability. 

\begin{figure}[t]
    \centering
    \includegraphics[width=\columnwidth]{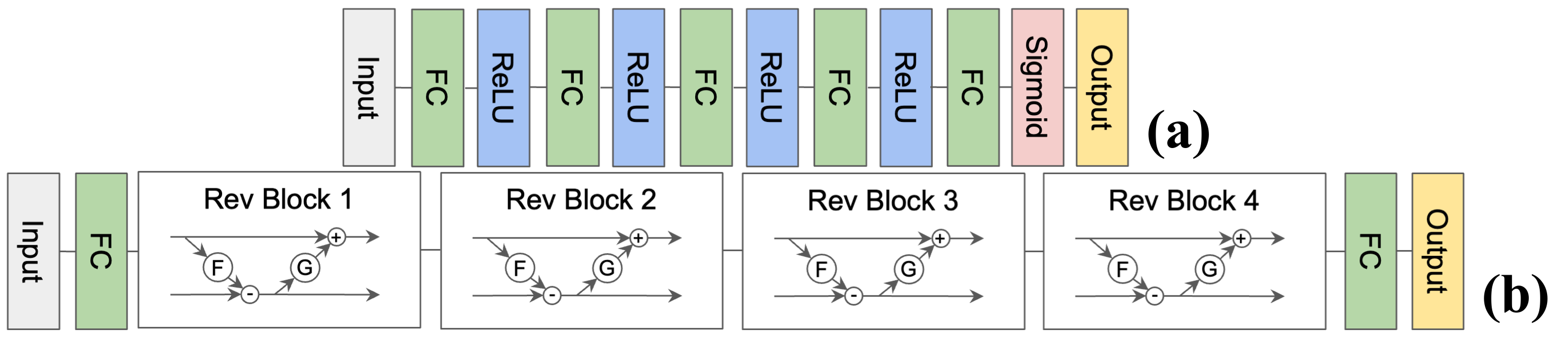}
    \caption{Model structure. \textbf{(a)} Workspace representation model using MLP. \textbf{(b)} Forward kinematics model using RevNet.}
    \label{fig:model}
    \vspace{-1.1cm}
\end{figure}

With the strong approximation capability of neural networks~\cite{devore2021neural}, to derive this complex nonlinear function $F_{w}$, we design a lightweight 6-layer MLP, as shown in Fig.~\ref{fig:model} (a). 
The three-dimensional coordinates are input to the network and transformed through five fully connected layers: first expanding to 64 neurons, then to 256 neurons, followed by reduction to 128 neurons, further reduction to 64 neurons, and finally outputting a single value. 
ReLU activation functions are employed between each fully connected layer, introducing nonlinear elements after each linear transformation, enabling the network to learn complex nonlinear mapping relationships. 
The final layer uses a Sigmoid activation function to compress the one-dimensional output value into the range of $[0,1]$, conforming to the range constraint of probability measures.

As demonstrated in Fig.~\ref{fig:represent} (b), the gradient of the output with respect to the input in deep neural networks which is $\frac{\partial \mathcal{P}}{\partial {}^{A}\boldsymbol{p}_e}$ can be obtained through backpropagation, and according to \cite{devore2021neural}, this process exhibits stability properties.
During the optimization process, this gradient guides points initially located outside the workspace towards higher-probability feasible regions as shown in Fig.~\ref{fig:represent} (c).


To quantitatively compare the workspace utilization efficiency between our method and that of \cite{cao2025motion}, we implemented a testing program using MATLAB. We reconstruct the delta workspace using the alpha shape method with $\alpha = 10$, compute its volume, and compare it with the inscribed cube volume. The analysis reveals that the volume of the inscribed cube is 551 cm$^3$, while our method achieves 2553 cm$^3$, representing a 363\% increase over the previous approach.
This substantial improvement in workspace utilization enables the delta arm to execute more flexible movements during aerial manipulation tasks.

\subsection{Optimization Framework}
In this work, we first employ the Jump Point Search (JPS) path-finding algorithm to generate a reference path that sequentially traverses all specified coordinate points.
Subsequently, we segment the path into $M$ sections and further improve the path quality using back-end optimization.

\subsubsection{Formulation}

To ensure the smoothness, feasibility of the trajectory, and its effective adaptation to manipulation task requirements, we construct the objective function of trajectory optimization as follows:
\begin{align}
\label{equ:object_function}
    &\mathcal{J}=\sum_{i=1}^M\mathcal{J}_{c,i}+\mathcal{J}_{t,i}\\
    &\text{s.t.} \quad \mathcal{G}(\boldsymbol{p}(t),\cdots,\boldsymbol{p}^{(4)}(t)) \le 0
\end{align}
where $\mathcal{J}_{c,i}$ is the control effort of the trajectory for the $i$-th segment, and $\mathcal{J}_{t,i}$ is the time regulation term.
The constraint function $\mathcal{G} \le 0$ ensures that the position $\boldsymbol{p}(t)$ and its derivatives up to the fourth order remain within their physical and operational limits.

We define $s=3$ for both the aerial manipulator and the end-effector trajectories. This choice of $s$ determines the form of the control effort term $\mathcal{J}_{c, i}$, which is given by:
\begin{equation}
\label{equ:control_effort}
    \mathcal{J}_{c,i} = \int_{0}^{T_i} \left \| \boldsymbol{p}_i^{(3)}(t) \right \|^2 dt
\end{equation}
where $\boldsymbol{p}_i^{(3)}(t)$ represents the jerk for the $i$-th trajectory segment. To balance trajectory smoothness with time efficiency, the time regulation term $\mathcal{J}_{t, i}$ aims to minimize the execution time of each segment:
\begin{equation}
\label{equ:time_regulation}
    \mathcal{J}_{t, i} = T_i
\end{equation}
where $T_i$ is the duration of the $i$-th trajectory segment.

To facilitate numerical optimization, the inequality constraint $\mathcal{G} \le 0$ can be transformed into a penalty term $\mathcal{J}_{p, i}$ integrated into the objective function. 
This approach handles constraint violations smoothly and is typically defined as:
\begin{equation}
\label{equ:penalty_term}
    \mathcal{J}_{p, i} = \sum_{u\in \mathcal{U}}w_u\mathcal{D}(\mathcal{J}_{\mathcal{G}_u,i})
\end{equation}
where $\mathcal{U}$ is the set of constraints, including dynamics and kinematics feasibility, collision avoidance constraints constructed with reference to~\cite{deng2025wholebody}, and task constraints discussed in Section~\ref{sec:task_constraints}.

The term $\mathcal{J}_{\mathcal{G}_{u, i}}$ represents the penalty function for the inequality constraint $u$ in the $i$-th trajectory segment, and
$w_u$ is the associated weight coefficient. The function $\mathcal{D}(\cdot)$ implements a sparse sampling strategy, which uses discrete numerical integration to approximate the continuous integral over the trajectory segment:
\begin{equation}
\label{equ:sparse_sample}
    \mathcal{D}(\mathcal{J}_{\mathcal{G}_u}) = \sum^{N-1}_{j=0}\mathcal{J}_{\mathcal{G}_{u,i}}(\frac{jT_i}{N})
\end{equation}
where $N$ represents the number of sampling points per trajectory segment, and $\frac{jT_i}{N}$ denotes the time at the $j$-th sampling point within the segment of duration $T_i$.


\subsubsection{Task Constraints}
\label{sec:task_constraints}
For manipulation tasks, the end-effector of the aerial manipulator is required to approach the operation point with a specific pose and complete the grasp task. To achieve this objective, we propose a set of task-specific constraints that ensure precise positioning and orientation planning.

Given the specific target positions $\boldsymbol{p}^s_{e, \gamma}$ and orientations $\boldsymbol{o}^s_{e, \gamma} \in \mathbb{R}^3$ of the end-effector in $\mathcal{F}_W$ , 
where $\gamma$ represents the index of the grasp phase in our manipulation sequence, we construct the following kinematic constraints to ensure precise object manipulation:
\begin{align}
\label{equ:pos_ori_constraint}
    \mathcal{J}_{\mathcal{G}_{K},i}& = 
    \begin{cases}
    \mathcal{S}(\mathcal{G}_{p,i})+\mathcal{S}(\mathcal{G}_{o,i}), & t= t_\gamma,\\
    0, &  \text{otherwise} .\\
    \end{cases}\\
\label{equ:penal_pos}
    \mathcal{G}_{p,i} &= \left \| \boldsymbol{p}_{e,i}(t_\gamma)-\boldsymbol{p}_{e,\gamma}^s \right \|^2  \\
\label{equ:penal_ori}
    \mathcal{G}_{o,i} &= \left \| \boldsymbol{f}_{i}(t_\gamma)- \| \boldsymbol{f}_{i}(t_\gamma) \| \boldsymbol{o}^s_{e, \gamma} \right \|^2 - e_o^2
\end{align}
where $t_\gamma = \sum^{l-1}_{j=0}T_j$ is the instant closest to the specific waypoint, $l \in \mathcal{I}$ denotes the index of the trajectory segment designated for executing the grasp actions, and $\mathcal{I}$ denotes the set of such segment indices.
$\mathcal{S}(\cdot)$ is the cubic smooth function given by $\mathcal{S}(x)= \text{max}(x^3, 0)$. 
$e_o$ is the maximum allowable angle difference, and $\boldsymbol{f}_{i}\in \mathbb{R}^3$ is the normalized thrust vector.

To enhance the motion stability of the end-effector at designated manipulation points, we also impose velocity constraints:
\begin{equation}
\label{equ:penal_vel}
    \mathcal{J}_{\mathcal{G}_{K},i} = 
    \begin{cases}
    0, & t = t_\gamma, \\
    \mathcal{S}(\| \dot{\boldsymbol{p}}_{e,i}(t_\gamma) \|^2), & \text{otherwise}.
    \end{cases}
\end{equation}

To guarantee the kinematic feasibility of the delta arm, we formulate workspace constraints using feasibility probability $\mathcal{P}$ mentioned in Sec.~\ref{sec:workspace_represent} to ensure the end-effector remains within the robot's reachable workspace:
\begin{equation}
\label{equ:penal_workspace}
    \mathcal{J}_{\mathcal{G}_{ws},i} = \sum^{N-1}_{j=0}\mathcal{S}(0.5 - \mathcal{P}_{i,j}).
\end{equation}
We establish 0.5 as the threshold for feasibility probability. 
When the model computes a feasibility probability greater than 0.5 for a given point, it indicates that the point likely resides within the workspace. 
Conversely, a probability less than 0.5 suggests that the point is distant from the workspace. 
Under this constraint, points evolve during the optimization process by moving toward regions of higher feasibility probability, effectively converging into the interior of the workspace.

Leveraging the gradient of the workspace representation model, we derive the gradient of the workspace penalty with respect to the end-effector position in $\mathcal{F}_A$:
\begin{equation}
\label{equ:workspace_gradient}
    \frac{\partial \mathcal{J}_{\mathcal{G}_{ws},i}}{\partial {}^{A}\boldsymbol{p}_{e,i,j}} = \frac{\partial \mathcal{J}_{\mathcal{G}_{ws},i}}{\partial \mathcal{P}_{i,j}} \frac{\partial \mathcal{P}_{i,j}}{\partial {}^{A}\boldsymbol{p}_{e,i,j}}.
\end{equation}

\subsection{Workspace Constraint Elimination}
In our framework, all trajectories can be parameterized by the time vector $\mathbf{T}$ and intermediate points $\mathbf{q}=[\boldsymbol{q}_1 \cdots \boldsymbol{q}_{M-1}]^T$ where $\boldsymbol{q}_i = [\boldsymbol{q}_{b, i}^T \quad  {}^{A}\boldsymbol{q}_{e, i}^T]^T$. 
These parameters are subject to specific constraints: $\boldsymbol{q}_{b, i}$ must lie within the polyhedrons, ${}^{A}\boldsymbol{q}_{e, i}$ must remain within the delta arm workspace, and all elements of the time vector $\mathbf{T}$ must be positive.
To handle the hard constraints on optimization variables, Wang \textit{et al.}~\cite{wang2022geometrically} proposed a polyhedron elimination method that transforms the constrained variables $\boldsymbol{q}_{b, i}$ and $\mathbf{T}$ into unconstrained variables.
However, this method is applicable only when the constraint space for each point is convex, making it unsuitable for handling the non-convex workspace constraints of ${}^{A}\boldsymbol{q}_{e, i}$ for the delta arm.
To address this limitation, we propose a learning-based workspace constraint elimination method specifically designed for the delta arm, which effectively transforms ${}^{A}\boldsymbol{q}_{e, i}$ into unconstrained variables while maintaining efficient gradient computation throughout the optimization process.

It is noteworthy that all the Cartesian coordinates within the delta arm's workspace can be independently computed through forward kinematics by servo joint angles $\boldsymbol{\theta}$ (Equation~\ref{equ:dynamics}).
To transform these joint angles into unconstrained variables for optimization, we define two variables $[\delta_{n},\zeta_n]$ for each joint angle where $n\in\{1,2,3\}$. For convenience, we normalize each joint angle to the range of $[0,1]$ using:
\begin{align}
\label{equ:normalization}
    \vartheta_n &= \frac{\delta_{n}^2}{\zeta_n^2+\delta_{n}^2}\\
    \theta_n &= 90 \cdot \vartheta_n
\end{align}
where $\vartheta_n \in [0,1]$ is the normalized value of the $n$-th joint angle.
Consequently, any end-effector coordinate within workspace, denoted as ${}^{A}\boldsymbol{q}_e$, can be determined by six unconstrained optimization variables:
\begin{equation}
\mathbf{\Xi} =[\delta_1 \quad \zeta_1 \quad \delta_2 \quad \zeta_2 \quad \delta_3 \quad \zeta_3]^T.    
\end{equation}

During the optimization process, updating $\mathbf{\Xi}$ through penalty terms requires gradient propagation, which is given by:
\begin{equation}
    \frac{\partial \mathcal{J}}{\partial \mathbf{\Xi}} = \frac{\partial \mathcal{J}}{\partial {}^{A}\boldsymbol{q}_e}\frac{\partial {}^{A}\boldsymbol{q}_e}{\partial \boldsymbol{\theta}}\frac{\partial \boldsymbol{\theta}}{\partial \boldsymbol{\vartheta}}\frac{\partial \boldsymbol{\vartheta}}{\partial \mathbf{\Xi}}.
\end{equation}

Due to the relatively complex forward kinematics of the delta arm, direct calculations of $\frac{\partial {}^{A}\boldsymbol{q}_e}{\partial \boldsymbol{\theta}}$ may lead to numerical instabilities, especially when the inverse kinematics is infeasible, thereby limiting its practical applicability.
To address this challenge, we propose using neural networks to approximate the forward kinematics, enabling stable gradient computation through the inherent differentiability of the network architecture.

To approximate the complex forward kinematics using the lightest possible network, we designed Delta RevNet (cf. Fig.~\ref{fig:model} (b)) with reference to RevNet~\cite{gomez2017reversible}. 
The model takes a three-dimensional vector with values in the range [0,1], which is $\boldsymbol{\vartheta}=[\vartheta_1 \quad\vartheta_2 \quad \vartheta_3]^T$ as input and ${}^{A}\boldsymbol{q}_e$ as output. 
The core of our network is the RevBlock (reversible block) structure. 
Each RevBlock evenly divides the input tensor along the feature dimension into two parts, then performs nonlinear transformations through two sequential networks, F and G, each composed of fully connected layers, batch normalization, and ReLU activation functions.

The complete Delta RevNet consists of an input projection layer, four RevBlocks, and an output projection layer. 
The input projection layer expands the original features into a higher-dimensional hidden space, ensuring an even dimension for block partitioning. 
The RevBlocks then sequentially perform reversible transformations, followed by the output projection layer, which maps the features back to the original dimension. 
This design can achieve better fitting effects with fewer parameters, ensuring that the network can rapidly compute output ${}^{A}\boldsymbol{q}_e$ through forward propagation and calculate gradients through backward propagation during the optimization process. 
Leveraging the automatic differentiators, we can get the model gradient $\frac{\partial {}^{A}\boldsymbol{q}_e}{\partial \boldsymbol{\vartheta}}$, which is equal to $\frac{\partial {}^{A}\boldsymbol{q}_e}{\partial \boldsymbol{\theta}}\frac{\partial \boldsymbol{\theta}}{\partial \boldsymbol{\vartheta}}$.

Finally, the simplified optimization problem is addressed through the L-BFGS method.

\begin{figure}[t]
    \centering
    \includegraphics[width=\columnwidth]{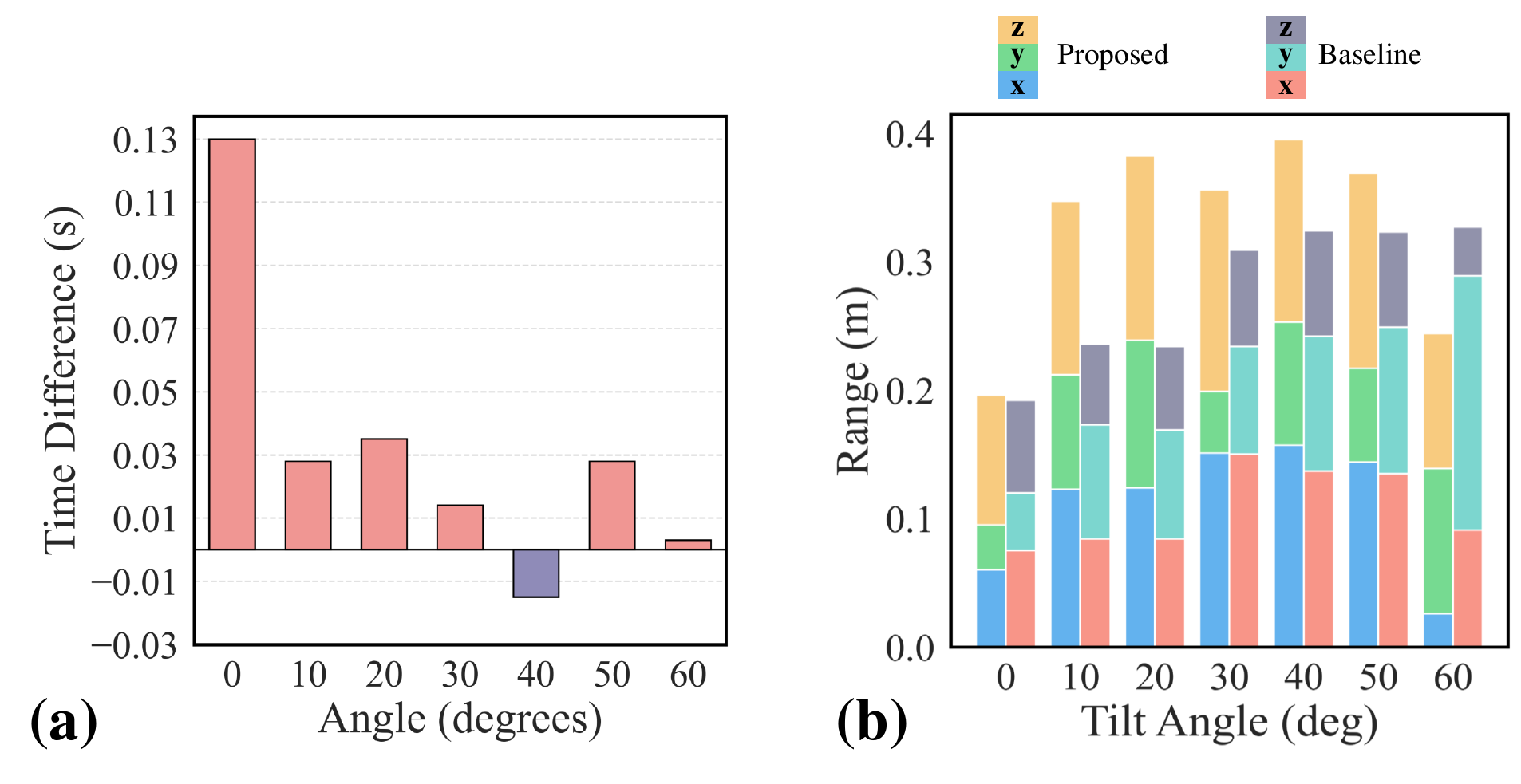}
    \caption{Comparison of the two methods in Scenario 1. \textbf{(a) }Difference in execution time $\Delta t = T_{\text{Baseline}} - T_{\text{Proposed}}$. \textbf{(b)} The maximum minus minimum end-effector position along each coordinate axis during execution.}
    \label{fig:s1data}
    \vspace{-0.68cm}
\end{figure}

\section{Experiments}

\subsection{Model Training}
The models used in all experiments are trained on a server with an NVIDIA RTX4090 GPU and CUDA 12.4. 
For the first model (cf. Fig.~\ref{fig:model} (a)), we generate the dataset by randomly sampling points within a bounded three-dimensional space and determining whether each point lies within the workspace through inverse kinematics calculations. 
Points outside the workspace are labeled as 0, while those inside are labeled as 1. 
For the second model (cf. Fig.~\ref{fig:model} (b)), we sample 3D vectors within the range $[0,1]$, denormalize them by scaling to 90, and compute the end-effector position through forward kinematics as labels. 
The two models are trained for 50 and 100 epochs, respectively, achieving excellent performance on the test set with error rates below 0.001\% across 100,000 test samples.

\subsection{Benchmark Comparison}
In this section, we compare the execution efficiency of our proposed planning framework with the method from~\cite{deng2025wholebody}, which substitutes the delta workspace with an inscribed cube. 
The comparison is performed in a simulated environment, where the aerial manipulator must position its end-effector at predefined target points while meeting specific orientation requirements. 
Comparisons are conducted in two scenarios: 1) inclined object grasping and 2) pick-and-place.

\subsubsection{Inclined object grasping}
This scenario features fixed starting and ending points, with a stationary target point for the end-effector, positioned at $[0,0,1]$ m. 
To evaluate the aerial manipulator's grasping performance under various attitude constraints, we established different inclination angles relative to the ground plane for the target point: $0^\circ,10^\circ,20^\circ,30^\circ,40^\circ,50^\circ,60^\circ$. 
When passing through the target point, the end-effector must reach the point and satisfy these angular requirements. 
We record the execution time and end-effector range of motion for both methods across different angles. 
The results are presented in Fig.~\ref{fig:s1data}.

As shown in Fig.~\ref{fig:s1data} (a), our method demonstrates significant advantages in execution efficiency, with substantially lower completion times compared to~\cite{deng2025wholebody}. 
Furthermore, due to the utilization of an expanded workspace, our approach enables a greater range of motion for the delta arm, with notably larger Z-axis activity scope in $\mathcal{F}_A$ than the baseline (cf. Fig. \ref{fig:s1data} (b)). 
This enhanced operational range contributes to the observed performance improvements.
\begin{figure}[t]
    \centering
    \includegraphics[width=0.98\columnwidth]{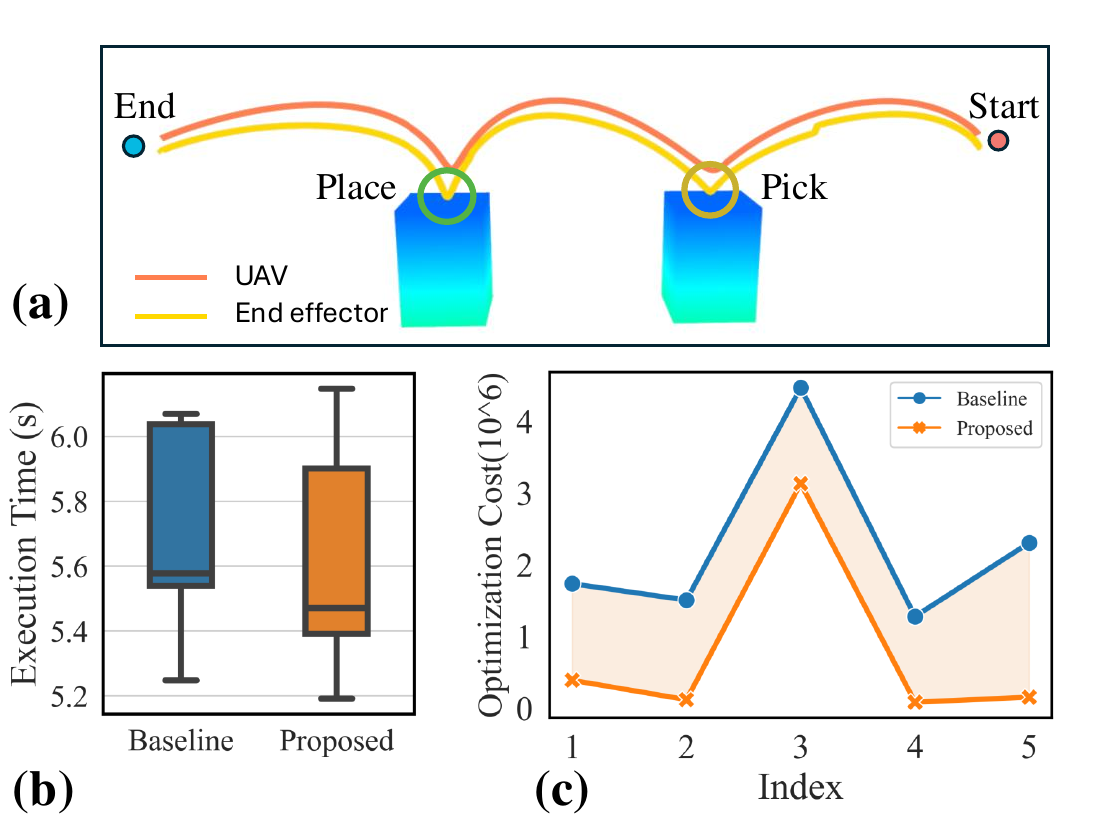}
    \caption{Simulation data in Scenario 2. \textbf{(a)} The overview of the pick-and-place scenario. \textbf{(b)} The execution time of the two methods. \textbf{(c)} The final optimization cost of the two methods.}
    \label{fig:s2data}
\end{figure}

\subsubsection{Pick-and-place}
In this scenario, we randomly generate two target points, each with an angular requirement of 0 degrees. 
The aerial manipulator is required to execute grasping at the first point, followed by a placement operation at the subsequent point, as illustrated in Fig.~\ref{fig:s2data} (a). 
We conducted five experiments using different random seeds, with the results presented in Fig.~\ref{fig:s2data} (b-c). Leveraging an expanded delta workspace, our proposed approach generates higher-quality trajectories, resulting in significantly reduced execution times. 
Moreover, our compact workspace constraint results in significantly reduced final costs (approximately 55\%) for the optimization problem, leading to more optimal solutions.

\begin{figure}[t]
    \centering
    \includegraphics[width=0.99\columnwidth]{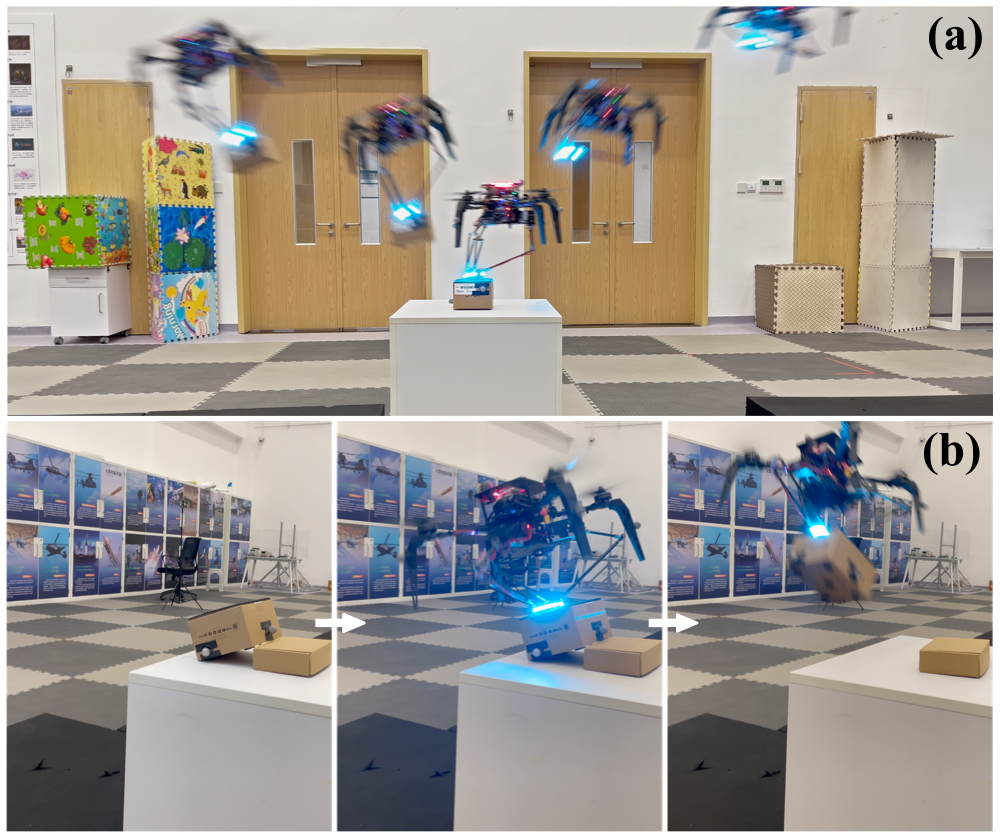}
    \caption{Snapshot of aerial grasping in real-world scenarios with different object placements. \textbf{(a)} Horizontally placed object. \textbf{(b)} Inclined object.}
    \label{fig:realworld}
    \vspace{-.68cm}
\end{figure}

\subsection{Real-World Experiments}
We further validated the feasibility of our method in real-world experiments, as shown in Fig.~\ref{fig:realworld}.
Our real-world platform consists of a quadrotor using NVIDIA Jetson Orin NX 16GB as the onboard computer and a delta arm with a hook-and-loop mechanism attached to its end-effector. 
The platform's state estimation is achieved by fusing data from the NOKOV Motion Capture System and the inertia measurement unit (IMU).
In this experiment, the aerial manipulator needs to execute an aerial grasping task, using the hook-and-loop mechanism to pick up rectangular boxes with corresponding hook-and-loop surfaces. 
The box is placed on a table of a certain height, and the experiment is divided into two scenarios (Experiment 1 and Experiment 2) based on different box placement angles. 
Due to the limited sensing capabilities of our experimental platform, we represented the scene using pre-constructed point cloud maps. The starting and ending points of the aerial manipulator remain fixed, while the position and orientation of the boxes are obtained through the NOKOV Motion Capture System. 
The controller design for the aerial manipulator follows the approach outlined in~\cite{chen2025ndob}.

In Experiment 1, the box was placed horizontally on the table surface, requiring the aerial manipulator's end-effector to approach the target with an orientation parallel to the table. 
For Experiment 2, the box was positioned at an inclined angle, necessitating the aerial manipulator to adjust its grasping orientation based on target pose information.

\begin{figure}[t]
    \centering
    \includegraphics[width=0.99\columnwidth]{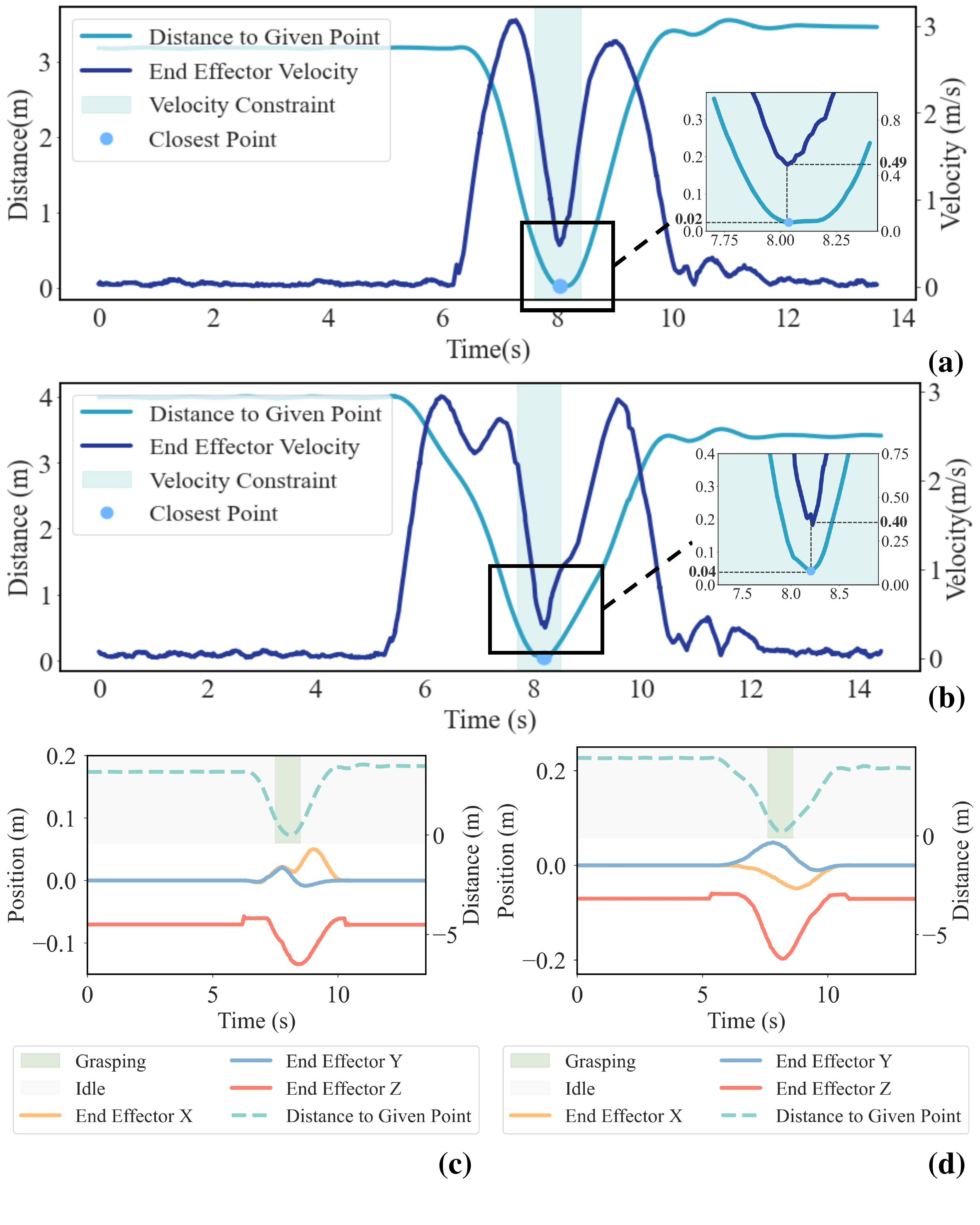}
    \caption{Flight data from real-world experiments. \textbf{(a-b)} Distance from the end-effector to the grasping point and its velocity in Experiments 1 and 2, respectively, with the velocity constraint active regions highlighted in light blue. \textbf{(c-d)} Three key pieces of information in Experiments 1 and 2, respectively: solid lines indicate the end-effector’s position along the x, y, and z axes; dashed lines represent the distance from the end-effector to the grasping point; and the aerial manipulator’s \textit{grasping} and \textit{idle} states are marked in light green and gray, respectively.}
    \label{fig:rwdata}
    \vspace{-0.38cm}
\end{figure}

As illustrated in Fig.~\ref{fig:realworld}, the aerial manipulator successfully completed the grasping tasks in both experiments, adhering to the target objects and transporting them away with speeds up to $3$ m/s. 
In Experiment 1, it performs high-speed grasping, picking up the object located $4$ meters away in just $1.8$ seconds.
Fig.~\ref{fig:rwdata} (a-b) depicts the end-effector velocity and its position relative to the target grasping point throughout the task execution of the two methods, respectively. 
When approaching the target point, the end-effector velocity rapidly decreases under the influence of velocity constraints in the task part, reaching $0.49$ m/s  and $0.40$ m/s at the grasping points respectively, ensuring stability during capture. 
Additionally, we assign relatively low weights to the velocity constraints to maintain reasonable operational efficiency. 
In terms of positioning accuracy, the end-effector achieves distances of $0.02$ m and $0.04$ m from the designated grasping points. 
These minimal errors ensured stable grasping while demonstrating the effectiveness of our framework.

Fig.~\ref{fig:rwdata} (c-d) displays the XYZ coordinates of the end-effector in $\mathcal{F}_A$ and its distance to the grasping point of the two methods, respectively. 
We define two states, \textit{grasping} and \textit{idle}, to distinguish between the aerial manipulator's grasping and non-grasping phases. 
During the initial \textit{idle} phase, the delta arm remains stationary to conserve energy. 
When the aerial manipulator approaches the target object, it transitions to the \textit{grasping} phase, where the arm extends significantly to facilitate rapid capture. 
After successful grasping, the delta arm moves to counteract the inertia of the captured object. 
Finally, it returns to the \textit{idle} state, with the arm retracted to minimize drag. 
These results demonstrate that with our thorough utilization of the delta workspace, the delta arm can assist in grasping through flexible movements.

\section{Conclusion}
In this work, we present a whole-body grasp planning framework for aerial manipulators equipped with delta arms.
We use a learning-based method to represent the complex delta arm workspace, maximizing its utility. 
Then, we formulate a coupled optimization problem to generate the trajectory for both the quadrotor and the delta arm, incorporating task constraints to ensure manipulation stability and effectiveness. 
We develop a learning-based approximation of the delta arm's forward kinematics, enabling efficient workspace constraint elimination. 
The framework's efficacy is validated through extensive simulations and real-world experiments. 
We show that our framework enables the UAV to grasp an object located 4 meters away within 2 seconds.
Future improvements for our framework might include enhancing real-time performance and autonomous exploration capabilities.
Although the proposed method successfully finds locally optimal feasible solutions, global optimality remains unguaranteed. Future research will explore methods to ensure global optimality.

\bibliography{RAL2018_reference} 
\end{document}